\documentclass{article}

\usepackage{arxiv}

\usepackage[utf8]{inputenc} 
\usepackage[T1]{fontenc}    
\usepackage{hyperref}       
\usepackage{url}            
\usepackage{booktabs}       
\usepackage{amsfonts}       
\usepackage{nicefrac}       
\usepackage{microtype}      
\usepackage{lipsum}
\usepackage{commands}
\usepackage{multirow}
\usepackage{graphicx}
\usepackage{amsmath}
\usepackage{tikz-cd}
\usepackage{mathtools}

\title{SeismoFlow - Data augmentation for the class imbalance problem}

\author{
  Ruy Luiz Milidiú \\
  Department of Computer Science\\
  Pontifical Catholic University of Rio de Janeiro\\
  Rio de Janeiro, Brazil \\
  \texttt{milidiu@inf.puc-rio.br} \\
   \And
 Luis Felipe Müller \\
  Department of Computer Science\\
  Pontifical Catholic University of Rio de Janeiro\\
  Rio de Janeiro, Brazil \\
  \texttt{lhenriques@inf.puc-rio.br} \\
}

\begin{document}
\maketitle

\begin{abstract}
    In several application areas, 
    such as medical diagnosis,  
    spam filtering, 
    fraud detection, 
    and seismic data analysis,
    it is very usual to find relevant classification tasks
    where some class occurrences are rare.
    This is the so called class imbalance problem,
    which is a challenge in machine learning.
    In this work, 
    we propose the {\ModelName} a flow-based generative model to create synthetic samples,
    aiming to address the class imbalance.
    Inspired by the Glow model, 
    it uses interpolation on the learned latent space to produce synthetic samples 
    for one rare class.
    We apply our approach to the development of a seismogram signal quality classifier.
    We introduce a dataset composed of
    {\DatasetSize} seismograms that are distributed between the 
        \textit{good}, 
        \textit{medium}, and 
        \textit{bad} 
    classes and with their respective frequencies of 
        {\GoodRatio}, 
        {\MediumRatio}, and 
        {\BadRatio}.
    Our methodology is evaluated on a stratified 10-fold cross-validation setting, 
    using the \textit{Miniception} model as a baseline, 
    and assessing the effects of adding the generated samples on the training set of each iteration.
    In our experiments, 
    we achieve an improvement of {\FScoreGain} on the rare class $F_{1}$ score, 
    while not hurting the metric value for the other classes and thus observing on the overall accuracy improvement.
    Our empirical findings indicate that our method can generate high-quality synthetic seismograms with realistic looking and sufficient plurality to help the \textit{Miniception} model to overcome the class imbalance problem.
    We believe that our results are a step forward in solving both the task of seismogram signal quality classification and class imbalance.
\end{abstract}


\section{Introduction} \label{sec:1-introduction}
    The class imbalance problem is a challenge in machine learning classification tasks \cite{Liu2009ExploratoryUF}. 
    The problem occurs when there is a 
    rare and very low-frequency class in the training set,
    making many machine learning algorithms, 
    such as neural networks, 
    struggle to learn to classify the low-frequency class \cite{Khan2015CostSensitiveLO}.
    Such a scenario appears in many different areas, 
    such as medical diagnosis,  
    spam filtering, 
    fraud detection, 
    and seismic data analysis.
    
    In this work, 
    we propose the {\ModelName} a flow-based generative model to create synthetic samples,
    aiming to address the class imbalance.
    Inspired by the Glow model \cite{Kingma2018GlowGF},  
    it uses interpolation on the learned latent space to produce synthetic samples for the rare class.
    We apply our approach to the development of a seismogram signal quality classifier.
    
    In geophysics, seismograms are used for a variety of tasks,
    such as to estimate subsurface structural settings, 
    stratigraphic geometry features,
    and potential hydrocarbon deposit locations \cite{chevitarese2018transfer}.
    Therefore, 
    one of the most important and time-consuming tasks to perform these analyses is the creation of a robust and credible dataset \cite{Valentine2010ApproachesTA}.
    
    Currently, 
    a huge amount of seismic data is routinely available, 
    but the quality of this data varies enormously.
    For example, 
    data from a seismic reflection surveying is usually corrupted by background noise,
    instrument malfunction, 
    or errors introduced during data storage and retrieval \cite{Valentine2010ApproachesTA}.
    This corrupted data may, 
    therefore, 
    lead to unusable seismograms. 
    Moreover, 
    for many applications, 
    the poor-quality data will distort results significantly, 
    which makes it impossible to use all available data blindly.
    Thus, 
    classifying the data according to its quality is mandatory when dealing with seismic data.
    
    We introduce a dataset composed of
    {\DatasetSize} seismograms that are distributed between the 
        \textit{good}, 
        \textit{medium}, and 
        \textit{bad} 
    classes and with their respective frequencies of 
        {\GoodRatio}, 
        {\MediumRatio}, and 
        {\BadRatio}.
    
    We evaluate our methodology on a stratified 10-fold cross-validation setting \cite{CV:Kohavi:1995}, 
    using the \textit{Miniception} model \cite{} as a baseline and assessing the effects of adding the generated samples on the training set of each iteration.
    In our experiments, 
    we achieve an improvement of {\FScoreGain} on the low-frequency class $F_{1}$ score, 
    while not hurting the metric value for the other classes and thus observing an overall accuracy improvement.
    Our empirical findings indicate that our method can generate high-quality synthetic samples with realistic looking and sufficient plurality to help a classifier to overcome the class imbalance problem.
    We believe that our results are a step forward in solving both the task of seismogram signal quality classification and class imbalance.
    
        The remaining of this paper is organized as follows. 
    In Section \ref{sec:2-related-work}, 
    we perform a brief review of the related works on generative normalizing flow models. 
    In section \ref{sec:3-background} we give a brief introduction to normalizing flow models and its loss formulation.  
    Section \ref{sec:4-model} describes the proposed {\ModelName} model, its layers and the normalization schemes added to the loss.
    Section \ref{sec:5-experiments} describes our experiments, the setups, and reports our findings.
    Finally, 
    section \ref{sec:6-conclusion} concludes with a short discussion section.

\section{Related Work} \label{sec:2-related-work}
    
    Currently, learning widely applicable generative models is an active area of research. 
    Many advances have been achieved, 
    mostly with likelihood-based methods \cite{Kingma2013AutoEncodingVB, Graves2013GeneratingSW, Dinh2014NICENI, Oord2016PixelRN, Dinh2016DensityEU, Kingma2018GlowGF} 
    and generative adversarial networks (GANs) \cite{Goodfellow2014GenerativeAN}.
     
    Likelihood-based models such as
    autoregressive models \cite{Hochreiter1997LongSM, Graves2013GeneratingSW, Oord2016ConditionalIG, Oord2016WaveNetAG}, 
    Variational autoencoders (VAEs) \cite{Kingma2013AutoEncodingVB} 
    and Flow-based generative models \cite{Dinh2014NICENI, Dinh2016DensityEU, Kingma2018GlowGF},
    learn from the optimization of a lower bound on the log-likelihood of the data.
    In contrast, 
    GANs \cite{Goodfellow2014GenerativeAN} learn in an adversarial training setting.
    In such a procedure,
    the generator network creates new data instances,
    while the discriminator network evaluates the samples for authenticity.
    GANs are known for their ability to synthesize large and realistic images \cite{Karras2017ProgressiveGO}.
    On the other hand, 
    the general lack of full support over the data points \cite{Grover2017FlowGANCM},
    the absence of encoders to the latent-space,
    the optimization instability,
    and the difficulty of evaluating overfitting and generalization,
    are some of the well-known disadvantages of GANs.
    
    In likelihood-based models,
    the variational autoencoder (VAE) algorithm \cite{Kingma2013AutoEncodingVB, Rezende2014StochasticBA} simultaneously learns a generative network
    and a matched approximate inference network.
    The generative network maps the Gaussian latent variables to samples.
    In its turn,
    by exploiting the reparametrization trick \cite{Williams1992SimpleSG},
    the approximate inference network maps samples to a semantically meaningful latent representation.
    The approximation in the inference process limits the model's ability to learn high dimensional deep representations.
    This limitation motivates a variety of works to attempt to improve the approximate inferences in variational autoencoders \cite{Maale2016AuxiliaryDG, Rezende2015VariationalIW, Salimans2014MarkovCM, Gibbs2000VariationalGP, Burda2015ImportanceWA, Kingma2016ImprovingVA}.
    
    Autoregressive models typically avoid such approximations by abstaining from using latent variables, 
    while usually retaining a great deal of flexibility \cite{Frey1998GraphicalMF, Larochelle2011TheNA, Germain2015MADEMA}.
    This class of algorithms simplifies the log-likelihood evaluation and sampling using the probability chain rule to decompose the joint distribution into a product of conditionals according to a fixed ordering over dimensions. 
    Usually, 
    works in this line of research use long-short term memory \cite{Hochreiter1997LongSM} recurrent networks,
    and residual networks \cite{He2016DeepRL, He2016IdentityMI} to learn generative image and language models \cite{Theis:2015:GIM, Oord2016PixelRN, Jzefowicz2016ExploringTL}.
    The ordering of the dimensions, 
    although often arbitrary, 
    can be critical to the training of the model \cite{Vinyals2015OrderMS}.
    Furthermore, 
    synthesis is difficult to parallelize, 
    and its computational effort is proportional to the data dimensionality \cite{Hochreiter1997LongSM, Graves2013GeneratingSW, Oord2016PixelRN, Oord2016ConditionalIG, Oord2016WaveNetAG, Oord2017ParallelWF}.
    Additionally, 
    there is no natural latent representation associated with autoregressive models.
    
    Finally,
    normalizing flow models \cite{Dinh2014NICENI, Dinh2016DensityEU, Kingma2018GlowGF},
    allow exact latent-variable inference and log-likelihood evaluation without approximation.
    The approximation absence leads to accurate inference, 
    and it also enables data direct log-likelihood optimization,
    instead of a lower bound of it \cite{Kingma2018GlowGF}.
    For this reason, 
    using such models, 
    data points can be directly represented in a latent space,
    making it easy to sample from the model, 
    perform semantic manipulations of the latent variables,
    interpolate data points,
    and evaluate the likelihood of a sample.
    Flow based models are usually built by stacking individual and simple invertible transformations that map the observed data to a standard Gaussian latent variable \cite{Dinh2014NICENI, Dinh2016DensityEU, Kingma2018GlowGF, Rezende2015VariationalIW, Kingma2016ImprovingVA, Louizos2017MultiplicativeNF, Ho2019FlowIF}.
    The number of stacked layers needed to map the observations to Gaussian space could be a limiting constraint to the model, 
    due to the extra memory needed to compute the gradients during training.
    But, 
    as the RevNet paper \cite{Gomez2017TheRR} showed that computing gradients in reversible neural networks requires a constant amount of memory, 
    instead of linear in their depth, 
    it does not become a problem in practice.
    
    This work builds upon the ideas and flows proposed in Nice \cite{Dinh2014NICENI}, RealNVP \cite{Dinh2016DensityEU}, and Glow \cite{Kingma2018GlowGF} to create synthetic augmentations of a low frequency class in an imbalanced dataset.

\section{Flow based Models} \label{sec:3-background}

    Let $x \in \mathbb{R}^M$ be a random vector with the true distribution $p(x)$. 
    Given an i.i.d. dataset $\mathcal{D}$ 
        and the discrete data $x \sim p(x)$,
        the log-likelihood objective is equivalent to minimizing:
            \begin{equation}
                \label{eq:data-log-likelihood-discrete}
                \mathcal{L}(\mathcal{D}) = \frac{1}{N} \sum^{N}_{i=1} -log\ p(x^{(i)})
            \end{equation}
    
    In the case of continuous data $x$,
        we first define
            $$
                \tilde{x}^{(i)} = x^{(i)} + u
            $$
        with $u \sim U(0,a)$,
        where $a$ is the data discretization level.
   Then, 
        we minimize
            \begin{equation}
                \label{eq:data-log-likelihood-continuous}
                \mathcal{L}(\mathcal{D}) \simeq \frac{1}{N} \sum^{N}_{i=1} -log\ p(\tilde{x}^{(i)}) + c
            \end{equation}
        where $c = -M \cdot log\ a$ and $M$ is the $x$ dimensionality.
    The loss defined in equations \ref{eq:data-log-likelihood-discrete} 
        and \ref{eq:data-log-likelihood-continuous} measures the expected compression cost in
        bits or 
        nats \cite{Dinh2016DensityEU}.
    
   The flow model
            \begin{equation}
                \label{eq:flow-model}
                z = f_{\theta}(x),
            \end{equation}
        where $\theta$ is the model's parameter collection,
        is constructed by composing a sequence of $L$ invertible flows
            \begin{equation}
                \label{eq:model-compositon}
                z = f_{\theta}(x) = f^{(1)} ( \dots ( f^{(L)}(x)),
            \end{equation}
       the so called normalizing flow \cite{Rezende2015VariationalIW}.
   Each $f^{(i)}$ component has a tractable inverse 
    and a tractable Jacobian determinant.
   Therefore, 
        the generative process is defined as:
            \begin{equation}
                \label{eq:z-follows-p}
                z \sim p_{z}(f_{\theta}(x))
            \end{equation}
            \begin{equation}
                \label{eq:x-fz}
                x = f_{\theta}^{-1}(z)
            \end{equation}
        where $p_{z}$ is typically a predefined simple and tractable distribution, 
        for example, 
        a standard Gaussian distribution.
    Thus,
        the function $f_{\theta}$ is bijective, 
        such that given a data point $x$ latent-variable inference is made by $z = f_{\theta}(x)$.
    
    The function $f_{\theta}$ is learned via maximum likelihood through the change of variable formula:
        \begin{equation}
            \label{eq:change-of-variable-formula}
            p(x) = p_{z}(f_{\theta}(x))\left | det \left (\frac{\partial f_{\theta}(x)}{\partial x} \right ) \right |
        \end{equation}
    Since $f_{\theta}$ is a composition of functions,
        the log probability density function given $z$, 
        under equation \ref{eq:change-of-variable-formula}, 
        is written as:
            \begin{equation}
                \label{eq:log-density}
                \begin{split}
                    log\ p(x) & = log\ p_{z}(f_{\theta}(x)) + log\ \left | det \left (\frac{\partial z}{\partial x} \right ) \right | \\
                                & = log\ p_{z}(f_{\theta}(x)) + \sum_{i=1}^{K} log\ \left | det \left (\frac{\partial h^{(i)}}{\partial h^{(i-1)}} \right ) \right |    
                \end{split}
            \end{equation}
        where $h^{(0)} = x$ is the observed data, 
        $h^{(i)}$ is the $i^{th}$ hidden layer output, 
        and $h^{(K)} = f_{\theta}(x) = z$ is the encoded latent variable. 
    Also,
        $\left |det(\partial h^{(i)}/ \partial h^{(i-1)}) \right |$ is the absolute value of the Jacobian determinant,
        and $log$ takes the logarithm of this value.
    The resulting scalar is also called the log-determinant. 
    It measures the change in log-density when going from $h^{(i-1)}$ to $h^{(i)}$ under the transformation $f^{(i)}$.
    
    Previous works 
    \cite{Rezende2015VariationalIW, Kingma2016ImprovingVA, Dinh2014NICENI, Dinh2016DensityEU}
    suggest using a family of simple transformations in which the Jacobian $\partial h^{(i)} /\partial h^{(i-1)}$ is a triangular matrix,
        making the log-determinant simple to calculate.
    \begin{equation}
        \label{eq:log-determinant}
        log \left | det\ \left( \frac{\partial h^{(i)}}{\partial h^{(i-1)}} \right) \right| = 
        sum \left( log \left| diag\ \left(\frac{\partial h^{(i)}}{\partial h^{(i-1)}}\right) \right| \right)
    \end{equation}
    In equation \ref{eq:log-determinant}, 
        $sum()$ takes the sum over all vector elements, 
        $log()$ takes the element-wise logarithm, 
        and $diag()$ takes the diagonal of the Jacobian matrix.
    
\section{SeismoFlow} \label{sec:4-model}
    In this section, 
    we describe the proposed {\ModelName} model. 
    {\ModelName} is a normalizing flow model, 
    inspired by NICE \cite{Dinh2014NICENI}, RealNVP \cite{Dinh2016DensityEU}, and Glow \cite{Kingma2018GlowGF} models. 
    Like in Glow,
    our model consists of a series of stacked flow blocks, 
    which each block is composed of the following sequence:
    \begin{figure}[h!]
        \centering
        \begin{tikzcd}
            \text{ActoNorm Layer} \arrow{r} & 
            \text{Invertible $1 \times 1$ Convolution} \arrow{r} & 
            \text{Coupling Layer}
        \end{tikzcd}
        \caption{Step Flow block.}
        \label{fig:step-flow-block}
    \end{figure}
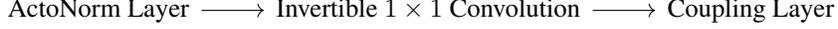
    
    In our model, we follow the design choice of \cite{Dinh2016DensityEU}, 
    and at regular intervals,
    we factor out half of the dimensions.
    This approach is known as multi-scale architecture and is described in subsection \ref{subsec:multi-scale}. 
    
    Unlike previous works on this kind of model, 
    we propose using of a multi-head self-attention layer \cite{Vaswani2017AttentionIA} within the coupling layers that belong to the two last scale levels.
    
    In the remainder of this section,
    we shortly explain each component that builds our model. 
    
    \subsection{ActNorm layer}
    Introduced in \cite{Kingma2018GlowGF},
    the Actnorm layer is proposed as an alternative to the standard batch normalization \cite{Ioffe2015BatchNA} for models trained with a small minibatch size.
    Similar to the batch normalization, 
    the Actnorm layer normalizes the activations per channel.
    This normalization is done by an affine transformation that uses the scale and bias learnable parameters.
    The scale and bias parameters are initialized such that the activations after the Actnorm layer have per-channel zero mean and unit variance given an initial minibatch of data.
    Finally, the log-determinant is computed as follows:
        \begin{equation}
            \label{eq:acnorm-logdet}
            log \left | det\left ( \frac{ \partial s \odot x + b}{\partial x} \right ) \right | = h \cdot w \cdot sum(log|s|)
        \end{equation}
    In equation \ref{eq:acnorm-logdet},
    $x$ is the layer input, 
    $s$ is the scale parameter, 
    $b$ is the bias parameter, 
    $\odot$ is an element-wise product operation, 
    $h$ and $w$ stand to the input height and width spatial dimensions.

    \subsection{Invertible $1\times 1$ convolution}

    The Glow model \cite{Kingma2018GlowGF} proposes to replace the usually used fixed permutations \cite{Dinh2014NICENI, Dinh2016DensityEU} by a learned invertible $1 \times 1$ convolution.
    The weight matrix of this convolutions is initialized as a random rotation matrix, 
    and they claim that since the number of channels in the layer's inputs and outputs is the same, 
    the  $1 \times 1$ convolution is a generalization of a permutation operator.
    
    The invertible $1 \times 1$ convolution receives a tensor $x \in \mathbb{R}^{h \times w \times c}$ as input and applies the convolutional operator using its weight matrix $W \in \mathbb{R}^{c \times c}$ as a filter.
    Additionally,  
    the weight matrix $W$ is initialized to have an initial log-determinant of $0$. 
    
    The log-determinant of this layer is computed as follows:
        \begin{equation}
            \label{eq:inv-conv-logdet}
            log \left | det\left ( \frac{\partial conv2D(h; W)}{\partial h} \right ) \right | = hight \cdot width \cdot log|det(W)|
        \end{equation}

    \subsection{Affine Coupling Layers}
    The affine coupling layer, 
    introduced in \cite{Dinh2014NICENI, Dinh2016DensityEU}, 
    is a simple bijective layer that is simple to invert and has a tractable Jacobian determinant. 
    The input vector is split into two halves, 
    and then complex dependencies are computed from one part and used to update the other part.
    In the end, 
    the layer concatenates the split parts into a vector containing the same dimensions as the input. 
    Given a $D$ dimensional input $x$ and $d < D$, 
    the output $y$ of an affine coupling layer follows the equation \ref{eq:affine-coupling-layer}
        \begin{equation}
            \label{eq:affine-coupling-layer}
            \left\{
            \begin{matrix}
                y_{1:d} = x_{1:d}; \\ 
                y_{d+1:D} = x_{d+1:D} \odot  exp(s(x_1:d)) + t(x_{1:d})
            \end{matrix}
            \right.
        \end{equation}
    where $s$ and $t$ stand for scale and translation 
    and are functions from $\mathcal{R}^d \rightarrow R$, 
    and $\odot $ is the element-wise product.
    
    The Jacobian of this transformation is
        \begin{equation}
            \label{eq:affine-coupling-jacobian}
            J =  
            \begin{bmatrix}
                \mathbb{I}_{d} & 0_{d\times(D-d)}\\ 
                \frac{\partial y_{d+1:D}}{\partial x_{1:d}} & diag(exp(s(x_{1:d})))
            \end{bmatrix}
        \end{equation}
    where $diag(exp(s(x_{1:d})))$ is the diagonal matrix whose diagonal elements correspond to the vector $s(x_{1:d}))$. 
    A crucial observation is that this Jacobian is upper triangular, 
    and therefore, 
    we can efficiently compute its determinant as $exp \left[ \sum_{j} s( x_{1:d})_{j} \right ]$.
    
    In the affine coupling layer, 
    the operators $s$ and $t$ can be any arbitrarily complex functions,
    such as a neural network, 
    because computing the Jacobian determinant of the layer does not involve calculating the Jacobian of those operators. 
    For convenience, in the remaining of this paper, we call the operators $s$ and $t$ as $NN()$.
    
    Finally, 
    computing the inverse of the coupling layer is no more complicated than the forward propagation, 
    and therefore, 
    sampling is as efficient as inference for this layer. 
    Equation \ref{eq:affine-coupling-layer-inverse} defines the inverse transformation of the affine coupling layer.
        \begin{equation}
            \label{eq:affine-coupling-layer-inverse}
            \left\{
            \begin{matrix}
                x_{1:d} = y_{1:d}; \\ 
                x_{d+1:D} = (y_{d+1:D} - t(y_{1:d})) \odot  exp(-s(y_{1:d}))
            \end{matrix}
            \right.
        \end{equation}
    
    \subsubsection*{Zero initialization} 
    Following \cite{Kingma2018GlowGF},
    we initialize the last $NN()$ layer with zeros.
    Thus,
    using this initialization, 
    each affine coupling layer initially performs an identity function. 
    
    \subsubsection*{Split and concatenation}
    The affine coupling layer splits the input tensor $h$ into two halves, 
    and at the end, 
    the layer undoes the split by concatenating the two halves into a single tensor again.
    Although other works suggest different types of splits, 
    such as splitting using a checkerboard pattern along the spatial dimensions \cite{Dinh2016DensityEU, Ho2019FlowIF}, 
    in this work, 
    we follow the strategy proposed by \cite{Dinh2014NICENI} and \cite{Kingma2018GlowGF}, 
    and only perform splits along the channel dimension. 
    \subsection{Multi-scale architecture} \label{subsec:multi-scale}

    Such as described in RealNVP \cite{Dinh2016DensityEU},
    we implement a multi-scale architecture using a squeezing operation.
    The squeezing operation divides the image into patches of shape $2 \times 2 \times c$,
    then reshapes them into patches of shape $1 \times 1 \times 4c$,
    transforming an $s \times s \times c$ tensor into an $\frac{s}{2} \times \frac{s}{2} \times 4c$ tensor.
    
    Our model is composed of a sequence of blocks, 
    which at regular intervals, 
    a squeeze operation is performed, 
    defining $L$ scale levels.
    At the beginning of each scale level, 
    we perform a squeeze operation, 
    and in the end, 
    we factor out half of the dimensions. 
    
    Thus,
    at each scale, 
    the spatial resolution is reduced and
    the number of hidden layer features is doubled.
    Additionally,
    all factored out variables are standardized to have zero mean and standard deviation one.
    The factored out variables correspond to intermediary representation levels that are fine-grained local features \cite{Rezende2014StochasticBA, Salakhutdinov2009DeepBM, Dinh2016DensityEU}.
    
    This approach reduces significantly the amount of computation and memory used by the model and distributes the loss function throughout the network.
    It is similar to guiding intermediate layers using intermediate classifiers \cite{Lee2014DeeplySupervisedN}.

\section{Experimental Setup and Results} \label{sec:5-experiments}
    
    In this section, 
    we present our experimental results, 
    a short description of the dataset,
    the experimental setup, 
    and the model hyper-parameters. 
    
    We start with the dataset description in sub-section \ref{subsec:5.1-dataset}; 
    secondly, 
        present the model architecture and hyper-parameters for both the {\ModelName} and the \textit{Miniception} models in sub-section \ref{subsec:5.2-setup};
    then, 
        we analyze the effects of varying the size of the augmentation on the classification performance in \ref{subsec:5.3-effects-of-augmentations};
    finally, 
        we close this section reporting the results for 
        the seismogram quality classification task measured in a cross-validation setting in sub-section \ref{subsec:5.4-cross-validation}.
    
\subsection{Dataset} \label{subsec:5.1-dataset}

\begin{figure}[ht]
        \label{fig:3-class-seismograms}
        \centering
        \includegraphics[width=12cm]{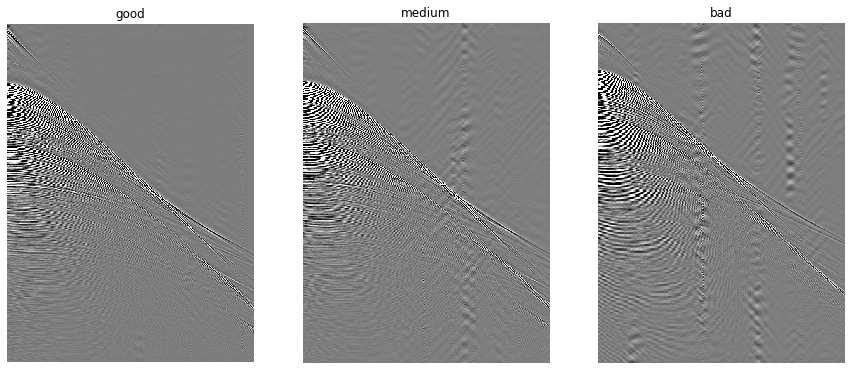}
        \caption{Per class seismogram example. In the left a sample from the good class. In the middle a sample from the medium class. In the left a sample from the bad class.}
    \end{figure}

    We conducted extensive experiments on highly diverse seismic data,
    and present our results on that dataset. 
    Our dataset comes from an offshore towed in a targeted
    region with $7993$ shot-gathers from $8$ cables each, 
    thus containing a total of $63994$ shot-gather images. 
    In a shot-gather image, each column corresponds to a seismic trace recorded during the same seismic shot. 
    Out of the total generated images, {\DatasetSize}  were randomly chosen and manually classified by a geophysicist with
        \textit{good}, 
        \textit{medium}, and 
        \textit{bad} 
    labels according to a visual inspection of artifacts related to swell-noise and anomalous recorded amplitude.
    The resulting dataset is composed of 
    {\DatasetSize} 
    seismograms that are distributed between the 
        \textit{good}, 
        \textit{medium}, and 
        \textit{bad} 
    classes and with their respective frequencies of 
        {\GoodRatio}, 
        {\MediumRatio}, and 
        {\BadRatio}.
    Figure \ref{fig:3-class-seismograms} shows one example per class of images.
    
\begin{figure}[ht]
        \label{fig:image-size-dist}
        \centering
        \includegraphics[width=8cm]{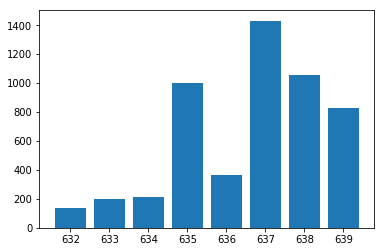}
        \caption{Distribution of images width sizes.}
    \end{figure}

    All images in our dataset have height resolution of $876$, 
    and the width varying between $[632,639]$.
    Figure \ref{fig:image-size-dist} shows the resolution distribution between dataset images.
    
    As it is explained in section \ref{sec:4-model}, our model gradually reduces the image resolution by two several times. Due to that architectural choice, in our experiments we fix the size of all images to a fixed resolution of {\DatasetResolution}. 
    
    
    In the following sections we explain our experimental setup and dataset splits for each of our experiments. Additionally, we present the obtained results and analysis. 
\subsection{Experimental Setup} \label{subsec:5.2-setup}
    Our models share the same settings and hyper-parameters across all experiments. 
    In the following, we present the settings for our models.
    
    \subsubsection*{Generative Model}
    We use multi-scale architecture with {\NLevels} scale levels 
    and {\NBlocks} step-flow blocks per level. 
    In the first four scale levels, 
    we let each coupling layer \textit{NN()} have three convolutional layers where the two hidden layers have ReLU activation functions and {\NFilters} channels. 
    The first and last convolutions are $3 \times 3$, 
    while the center convolution is $1 \times 1$. 
    Additionally, 
    before the last convolution, 
    we apply a layer normalization \cite{Ba2016LayerN}.
    
    In its turn,
    in the last scale level,
    we let each coupling layer \textit{NN()} have one 4-headed self-attention layer \cite{Vaswani2017AttentionIA}, one layer normalization, $3 \times 3$ convolution layer with ReLu activation, and one $3 \times 3$  convolutional layers with its weights initialized with zeros, 
    all with {\NFilters} filters. 
    
    Our model is trained using Adam Optimizer \cite{Kingma2014AdamAM}, 
    with the default hyper-parameters, 
    during {\NumberOfEpochs} epochs. 
    Due to memory constraints, 
    we use a batch size of {\BatchSize} image per iteration step. 
    Additionally, 
    the learning rate is scheduled to decay in a polynomial shape with a warm-up phase of {\WarmUpSteps} steps and a maximum learning rate of {\LearningRate}.

\subsubsection*{Classification Model}
    We reuse all settings for the Miniception model \cite{miniception2019deep},
    including hidden sizes and initialization.
    
    The model is trained using Adam Optimizer \cite{Kingma2014AdamAM}, 
    with the default hyper-parameters for {\NEpochs} epochs.

\subsection{Augmentations Effects} \label{subsec:5.3-effects-of-augmentations}

    We begin our experiments by analyzing the effects of our generated samples on the \textit{Miniception} model performance on the task of classifying the signal quality of seismograms.
    For this experiment,
    we performed the usual division of 
        training,
        validation,
        and testing on our dataset 
    using 
        {\TrainPartitionPercent},
        {\ValidPartitionPercent}, 
        and {\TestPartitionPercent} 
    of the data for each respective partition.
    The partitions are stratified to maintain the original dataset distribution of classes.
    Additionally, 
    the model is trained Adam Optimizer \cite{Kingma2014AdamAM},
    with the default hyper-parameters and early stop with five epochs of patience for at maximum $100$ epochs.
    
\begin{figure}[ht]
        \label{fig:aug_performance_curve}
        \centering
        \includegraphics[width=8cm]{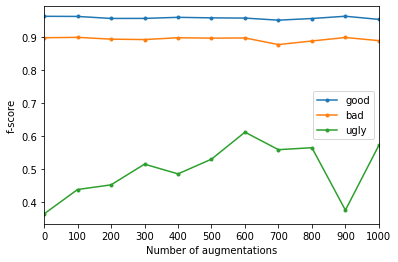}
        \caption{$F_{1}$ score curve when varying the augmentation size.}
    \end{figure}

    We present the performance curve of the $F_{1}$ metric, 
    across ten runs of the learning process,
    using different augmentations sizes for low-frequency class.
    Figure \ref{fig:aug_performance_curve} show the evolution of the $F_{1}$ score when varying the number of augmentations. The evaluation starts with zero augmentations and ends with $1000$ augmentations for the minority class.
    
    Since the $F_{1}$ score captures the balance between the $Precision$ and $Recall$, 
    tending to the lowest value,
    accessing the figure above,
    we can conclude that for this set of partitions,
    {\NSamples} is the number of augmentations that achieves the most significant balanced gain in the metrics for all classes.
    
    We believe that performing this search with cross-validation would give us a more robust result, 
    but the execution time and computational power required for such an experiment prevents us from performing it.
    Thus, 
    we use this result as a hint to the model's expected behavior to define the hyper-parameters of the next experiment that is presented in the following subsection.
    
\subsection{Cross-Validation} \label{subsec:5.4-cross-validation}
    In this section,
    we evaluate the quality of our methodology by augmenting the training set of our dataset using the samples generated by our generative model.
    We evaluate the $Precision$, 
    $Recall$, 
    and $F_{1}$ metrics,
    and its deviations,
    in a stratified 10-fold cross-validation setting \cite{CV:Kohavi:1995}.
    In this setting, 
    we split our data in 10 non-overlapping folds that are per-class stratified, 
    fitting the \textit{Miniception} model in 9 folds and leaving one out to test.
    In all cross-validation iterations,
    the \textit{Miniception} model is trained for {\NEpochs} epochs using the setup explained in section \ref{subsec:5.2-setup}.
    
    In Table \ref{tab:metrics},
    we present the obtained results, 
    with and without augmenting the training set with our generated samples. 
    For the augmented version,
    we generate {\NSamples} samples using the {\ModelName} model for each cross-validation iteration.
    Additionally, 
    to avoid passing any information from the test folds to the training folds,
    the augmentations are generated by only using data from the training set of each cross-validation iteration.

    \begin{table}[h]
            \caption{
                Cross-Validation per-class average 
                precision, 
                recall, 
                and $F_{1}$ metrics.
            }
            \centering
            \begin{tabular}{l|lll|lll}
                \toprule
                 \multirow{2}{4em}{Class} & 
                 \multicolumn{3}{c}{Not Augmented} & 
                 \multicolumn{3}{c}{Augmented} \\
                 
                 &  $Precision$ (\%) & 
                    $Recall$ (\%) & 
                    $F_{1}$ (\%) &  
                    $Precision$ (\%) & 
                    $Recall$ (\%) & 
                    $F_{1}$ (\%) \\
                
                \midrule
                
                Good    & 
                    {\GoodNoAugPrecision}  & 
                    \textbf{\GoodNoAugRecall} & 
                    {\GoodNoAugFScore} & 
                    \textbf{\GoodAugPrecision}  & 
                    {\GoodAugRecall} & 
                    \textbf{\GoodAugFScore} \\
                         
                Medium & 
                    {\BadNoAugPrecision} & 
                    {\BadNoAugRecall} & 
                    {\BadNoAugFScore} &
                    \textbf{\BadAugPrecision} & 
                    \textbf{\BadAugRecall} & 
                    \textbf{\BadAugFScore} \\
                        
                Bad    & 
                    {\UglyNoAugPrecision}  & 
                    {\UglyNoAugRecall} & 
                    {\UglyNoAugFScore} &
                    \textbf{\UglyAugPrecision}  & 
                    \textbf{\UglyAugRecall} & 
                    \textbf{\UglyAugFScore}\\
                \bottomrule
            \end{tabular}
            \label{tab:metrics}
        \end{table}
        
    As can be seen from the table, 
    our augmentations yield consistent improvements in metrics,
    not only for the bad class but also for the other classes.
    We can note that when adding the augmentations, 
    the model improves
        {\UglyPrecisionGain} in the precision,
        {\UglyRecallGain} in the recall,
        and {\UglyFScoreGain} in the $F_{1}$ scores of the low-frequency class. 
    Furthermore, we can observe these improvements as long as the values for these metrics on the other classes are not impaired and thus, improving the model's overall accuracy.
    
    Table \ref{tab:averaged-metrics} show averaged metrics for both versions of the model, with and without data augmentation.
    We can note that, when augmenting the training data, the model gets a {\FScoreGain} improvement on the averaged $F_{1}$ metric.
    
    \begin{table}[h]
            \caption{
                Cross-Validation average 
                precision, 
                recall, 
                and $F_{1}$ metrics.
            }
            \centering
            \begin{tabular}{l|lll}
                \toprule
                 Model & 
                 $Precision$ (\%) & 
                 $Recall$ (\%) & 
                 $F_{1}$ (\%) \\
                \midrule
                
                Augmented    & 
                    \textbf{\PrecisionAug}  & 
                    \textbf{\RecallAug} & 
                    \textbf{\FScoreAug} \\

                Not Augmented & 
                    {\PrecisionNoAug} & 
                    {\RecallNoAug} & 
                    {\FScoreNoAug} \\
                \bottomrule
            \end{tabular}
            \label{tab:averaged-metrics}
        \end{table}
    
\section{Conclusion} \label{sec:6-conclusion}
    
    A common problem in machine learning classification tasks
    is the imbalanced class problem \cite{Liu2009ExploratoryUF}. 
    The class imbalance problem is a challenge in machine learning classification tasks. The problem occurs when there is a 
    rare and very low-frequency class in the training set,
    making many machine learning algorithms, 
    such as neural networks, 
    struggle to learn to classify the low-frequency class correctly \cite{Khan2015CostSensitiveLO}.
    
    In this work, 
    we propose a flow-based generative model to create synthetic samples,
    aiming to address the class imbalance.
    Inspired by the Glow model \cite{Kingma2018GlowGF},  
    it uses interpolation on the learned latent space to produce synthetic samples for the smallest class.
    We apply our approach to the development of a seismogram signal quality classifier.
    
    We introduce a dataset composed of
    {\DatasetSize} seismograms that are distributed between the 
        \textit{good}, 
        \textit{medium}, and 
        \textit{bad} 
    classes and with their respective frequencies of 
        {\GoodRatio}, 
        {\MediumRatio}, and 
        {\BadRatio}.
    
    We evaluate our methodology on a stratified 10-fold cross-validation setting \cite{CV:Kohavi:1995}, 
    using the \textit{Miniception} model \cite{} as a baseline and assessing the effects of adding the generated samples on the training set of each iteration.
    
    To avoid transferring from the test fold set to the training folds set, 
    when augmenting the minority class, 
    we compute the augmentations,  
    only using the data designed to the training set of each cross-validation iteration.
    In our experiments, we analyze the effects of adding the generated seismograms by comparing 
    the \textit{precision}, 
    \textit{recall}, 
    and \textit{$F_{1}$} metrics of each class.
    When augmenting the training set with our generated samples,
    we achieve an improvement of {\FScoreGain} on the $F_{1}$ score of the low-frequency class, 
    while not hurting the metric value for the other classes and thus observing an improvement on the overall accuracy.
    Finally, 
    our empirical findings indicate that our method can generate high-quality synthetic samples with realistic looking and sufficient plurality to help a second model to overcome the class imbalance problem.
    We believe that our results are a step forward in solving both the task of seismogram signal quality classification and class imbalance.

\section*{Acknowledgment}
This material is based upon work supported by PETROBRAS, CENPES, and by the Air Force Office Scientific Research under award number FA9550-19-1-0020.
Also, this study was financed in part by the Coordenação de Aperfeiçoamento de Pessoal de Nível Superior – Brasil (CAPES) – Finance Code 001.

\bibliographystyle{unsrt}  
\bibliography{references}  
\end{document}